\pdfoutput=1
\documentclass[11pt]{article}

\usepackage{acl}

\usepackage{microtype}
\usepackage{times}
\usepackage{latexsym}
\usepackage{soul}
\usepackage{amsmath}
\usepackage{booktabs}
\usepackage{bm}
\usepackage{cleveref}
\usepackage{graphicx}
\usepackage{tabularx}
\usepackage{soul}
\usepackage{xcolor}
\usepackage{todonotes}
\usepackage{multirow}
\usepackage{xpatch}
\usepackage{blindtext}
\usepackage{fdsymbol}
\usepackage{microtype}
\usepackage{hyperref}
\usepackage{adjustbox}
\usepackage{textcomp}
\usepackage{xparse}
\usepackage{enumitem}
\usepackage{makecell}
\usepackage{subcaption}
\usepackage{colortbl}
\usepackage{array}

\crefformat{section}{\S#2#1#3} % see manual of cleveref, section 8.2.1
\crefformat{subsection}{\S#2#1#3}
\crefformat{subsubsection}{\S#2#1#3}

\usepackage[T1]{fontenc}
\usepackage[utf8]{inputenc}

\usepackage{microtype}

\usepackage{xcolor}

\definecolor{c2}{RGB}{218,0,0}

\definecolor{lightblue}{RGB}{212, 235, 255}
\definecolor{lightorange}{RGB}{255, 204, 168}
\definecolor{lightyellow}{RGB}{255, 255, 168}
\definecolor{lightred}{RGB}{255, 168, 168}
\definecolor{lightpink}{RGB}{255,204,204}
\definecolor{darkred}{RGB}{196, 30, 58}
\definecolor{lightgreen}{RGB}{213, 232, 212}
\definecolor{lightgreen}{rgb}{0.82, 0.94, 0.75}
\definecolor{lightpurple}{RGB}{225, 213, 231}
\definecolor{normalgreen}{rgb}{0.66, 0.9, 0.5}
\definecolor{lightgray}{rgb}{0.7, 0.7, 0.7}
\definecolor{gold}{RGB}{255,217,102}

\newcommand\hlc[2]{\sethlcolor{#1} \hl{#2}}

\newcommand{\data}[1]{\textsc{VALOR-Bench}}

\newcommand{\eval}[1]{\textsc{VALOR-Eval}}

\usepackage{pifont}% http://ctan.org/pkg/pifont

\definecolor{darkgreen}{rgb}{0.0,0.5,0.0}
\newcommand{\cmark}{\textcolor{darkgreen}{\ding{51}}}
\newcommand{\xmark}{\textcolor{red}{\ding{55}}}
\definecolor{mygray}{gray}{.90}
\newcommand{\colorgray}{\cellcolor{mygray}}

\definecolor{Green6}{rgb}{0.251, 0.753, 0.341}
\definecolor{blue-violet}{rgb}{0.54, 0.17, 0.89}

\usepackage{tcolorbox}

\begin{document}

\title{\eval{}: Holistic Coverage and Faithfulness Evaluation of \\ Large Vision-Language Models} %Multi-Dimensional Evaluation of Large Vision Language Models for Coverage-Faithfulness Tradeoff}

% \title{Coverage-Faithfulness Evaluation of Large Vision Language Models \\ through Multi-Dimensional Benchmarking}

%Deciphering the Mirage: \\ An LLM-based Coverage-Faithfulness Evaluation of Hallucinations in Large Vision-Language Models through Multi-Dimensional Benchmarking

% Coverage-Faithfulness Evaluation of Hallucinations in Large Vision-Language Models

% Author information can be set in various styles:
% For several authors from the same institution:
% \author{Author 1 \and ... \and Author n \\
%         Address line \\ ... \\ Address line}
% if the names do not fit well on one line use
%         Author 1 \\ {\bf Author 2} \\ ... \\ {\bf Author n} \\
% For authors from different institutions:
% \author{Author 1 \\ Address line \\  ... \\ Address line
%         \And  ... \And
%         Author n \\ Address line \\ ... \\ Address line}
% To start a seperate ``row'' of authors use \AND, as in
% \author{Author 1 \\ Address line \\  ... \\ Address line
%         \AND
%         Author 2 \\ Address line \\ ... \\ Address line \And
%         Author 3 \\ Address line \\ ... \\ Address line}

\author{Haoyi Qiu$^*$ ~~Wenbo Hu$^*$ ~~Zi-Yi Dou ~~Nanyun Peng \\ University of California, Los Angeles \\ \texttt{\{haoyiqiu,whu,zdou,violetpeng\}@cs.ucla.edu}}

\maketitle
{\def\thefootnote{*}\footnotetext{The authors contributed equally to this work and are listed in alphabetical order by first name.}}

\begin{abstract}

Large Vision-Language Models (LVLMs) suffer from \textit{hallucination} issues, wherein the models generate plausible-sounding but factually incorrect outputs, undermining their \textit{reliability}. A comprehensive quantitative evaluation is necessary to identify and understand the extent of hallucinations in these models. However, existing benchmarks are often limited in scope, focusing mainly on object hallucinations. Furthermore, current evaluation methods struggle to effectively address the subtle semantic distinctions between model outputs and reference data, as well as the balance between hallucination and informativeness. To address these issues, we introduce a multi-dimensional benchmark covering objects, attributes, and relations, with challenging images selected based on associative biases. Moreover, we propose a large language model (LLM)-based two-stage evaluation framework that generalizes the popular CHAIR metric~\cite{rohrbach-etal-2018-object} and incorporates both faithfulness and coverage into the evaluation. Experiments on 10 established LVLMs demonstrate that our evaluation metric is more comprehensive and better correlated with humans than existing work when evaluating on our challenging human-annotated benchmark dataset. Our work also highlights the critical balance between \textit{faithfulness} and \textit{coverage} of model outputs, and encourages future works to address hallucinations in LVLMs while keeping their outputs \textit{informative}.\footnote{Our dataset and code can be found here: \url{https://github.com/haoyiq114/VALOR}.}
   
\end{abstract}
\section{Introduction}
\label{sec:intro}

\begin{figure}[t]
 \centering
 \includegraphics[width=\linewidth]{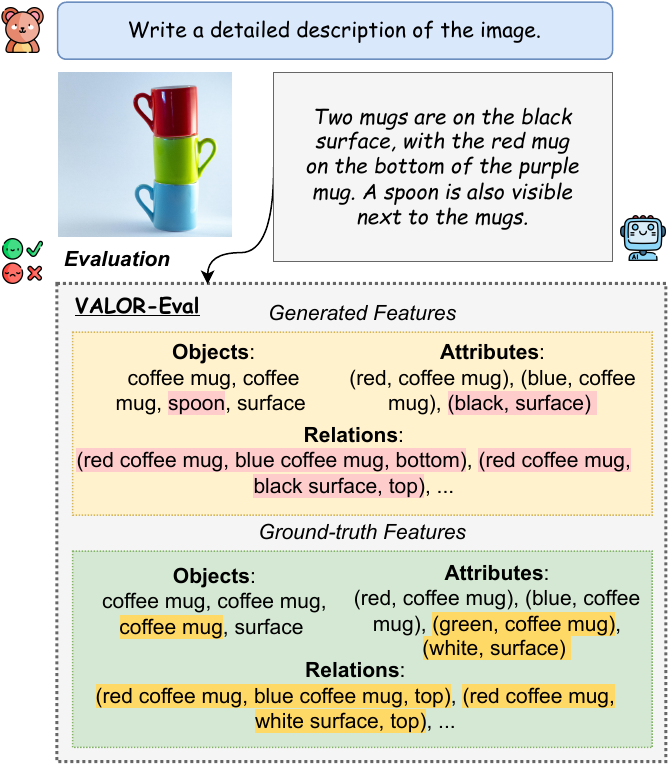}
 \vspace{-7mm}
 \caption{Example of the hallucination in open vocabulary generation task of LVLMs. Our proposed framework can identify objects, attributes, and relations from the generated captions and provide a comprehensive evaluation of faithfulness and coverage. We highlight \hlc{lightpink}{hallucinated} features and \hlc{gold}{uncovered} features.}
 \label{fig:intro}
 \vspace{-5mm}
\end{figure}

Large Vision-Language Models (LVLMs)~\cite{liu2023llava,Achiam2023GPT4TR,Chen2023MiniGPTv2LL} have shown remarkable performance across a broad range of vision-language tasks.
Despite the promising progress, the issue of hallucinations has emerged as a critical concern. \textit{Hallucination} refers to the generation of plausible-sounding but inaccurate or fabricated textual descriptions for a given image, which can compromise the reliability and trustworthiness of the models.

\begin{table*}[ht]
\small
\centering
\setlength{\tabcolsep}{3mm}{
\scalebox{0.93}{
\begin{tabular}{lccccccc}
\toprule
 \textbf{Evaluation} &\multicolumn{3}{c}{\textbf{Hallucination Type}}&\textbf{Human} & \multirow{2}{*}{\textbf{Faithfulness}} & \multirow{2}{*}{\textbf{Coverage}} & \textbf{Open Vocab.} \\
  \cmidrule(lr){2-4}  
 \textbf{Method} & \textbf{\textit{Object}} & \textbf{\textit{Attribute}} & \textbf{\textit{Relation}}  &\textbf{Annotation} &&& \textbf{Generation} \\
\midrule
POPE & \cmark & \xmark & \xmark & \xmark & \cmark & \xmark & \xmark\\
HaELM &\cmark &\textbf{?} &\textbf{?} & \xmark & \cmark & \xmark & \cmark\\ 
HallusionBench &\cmark &\textbf{?} &\textbf{?}  & \cmark  & \cmark & \xmark & \xmark \\ 
Halle-Switch &\cmark & \xmark&\xmark&\xmark & \cmark & \cmark & \cmark\\
NOPE &\cmark & \xmark&\xmark&\xmark  & \cmark & \xmark & \xmark \\
Bingo & \textbf{?} &\textbf{?}  &\textbf{?} &\textbf{?}  & \cmark & \xmark & \xmark  \\ 
FaithScore & \cmark &\cmark &\cmark &\xmark & \cmark & \xmark & \cmark \\
AMBER & \cmark& \cmark & \cmark& \cmark   & \cmark & \cmark & \xmark  \\
MERLIM & \cmark & \xmark &\xmark & \xmark & \cmark & \xmark & \xmark \\
\colorgray \textbf{Ours (\eval~)} & \colorgray \cmark & \colorgray \cmark & \colorgray \cmark & \colorgray \cmark & \colorgray  \cmark & \colorgray  \cmark &\colorgray  \cmark  \\

\bottomrule
\end{tabular}
}
}
\vspace{-3mm}
\caption{Comparison of existing hallucination evaluation benchmarks for LVLMs, including POPE \cite{li-etal-2023-evaluating}, HaELM \cite{Wang2023EvaluationAA}, HallusionBench \cite{guan2023hallusionbench}, Halle-Switch \cite{Zhai2023HallESwitchRA}, NOPE \cite{Lovenia2023NegativeOP}, Bingo \cite{cui2023holistic}, FaithScore \cite{faithscore}, AMBER \cite{wang2023llm}, MERLIM \cite{Villa2023BehindTM}. \textbf{?} refers to features not explicitly mentioned in the paper. Open Vocab represents evaluating free-form generated captions without constraints to pre-defined vocabulary.}
\vspace{-4mm}
\label{tab:evaluation_frameworks}
\end{table*}

Recent studies have proposed various methods to \textit{evaluate} models' \textit{generative} hallucinations \cite{Wang2023EvaluationAA,Zhai2023HallESwitchRA,faithscore} and \textit{discriminative} hallucinations \cite{li-etal-2023-evaluating,guan2023hallusionbench,Lovenia2023NegativeOP}. However, they predominantly focus on hallucinations concerning object existence and their faithfulness within generated content, often neglecting other critical types of hallucinations and the assessment of coverage. This oversight can result in a lack of attention to the variety and depth of hallucinations that may occur beyond object identification, such as attributes and relations. Furthermore, these evaluation methods are often constrained by a predefined vocabulary, thus are inherently limited to fully appreciating the richness of the free-form generated captions. Specifically, the evaluation metrics may not capture novel expressions that extend beyond the predetermined vocabulary. 

In contrast to prior studies, we introduce a human-annotated multi-dimensional evaluation benchmark \data~\footnote{VALOR is short for \underline{v}ision-language \underline{a}ttribute, re\underline{l}ation, and \underline{o}bject cove\underline{r}age and faithfulness.} by breaking down hallucinations into three categories: \textit{object} (existence), \textit{attributes} (color and count), and \textit{relations} (positional and comparative). In addition, to make the test cases challenging, we utilize the \textit{associative biases} \cite{li-etal-2023-evaluating,zhou2023analyzing} presented in training datasets to select images with only one component of commonly co-occurred pairs or groups, leading models to mistakenly generate associated elements that are not present. Our experimental findings validate the effectiveness of this methodology in exposing the susceptibility of current LVLMs to such biases.

In addition to constructing the benchmark dataset, we also propose a new evaluation framework, \eval{}. Existing evaluation frameworks such as the widely used CHAIR \cite{rohrbach-etal-2018-object} metric, exhibit several major constraints. First, they rely on a predefined vocabulary, limiting their ability to identify hallucinations in an \textit{open vocabulary} setting where semantic nuances -- such as synonyms and variations --  are prevalent in model outputs and references. Additionally, focusing exclusively on hallucination overlooking the aspect of \textit{coverage}, resulting in a preference for precise but uninformative model outputs. To address these issues, our propose \eval~ metric generalizes CHAIR by incorporating an LLM in a two-stage design, enhancing the capability to evaluate open vocabulary hallucination across object, attribute, and relation dimensions while also considering coverage. We provide a detailed comparison of existing evaluation methods in \Cref{tab:evaluation_frameworks}.

We conduct comprehensive evaluations on 10 established LVLMs across multiple dimensions with \data~. Our findings reveal that some LVLMs tend to prioritize precision over coverage, leading to predictions with high accuracy but limited scope. This observation underscores the need for the community to focus on achieving an \textit{balance} between faithfulness and coverage in LVLMs. Our contributions are threefold:

\begin{itemize}
    \item We introduce \data~, a comprehensive human-annotated dataset covering \textit{relation}, \textit{attribute}, \textit{object} with challenging images selected based on associative bias.
    \item We propose an LLM-based two-stage evaluation framework \eval~ that generalizes previous methods to consider the precision and informativeness trade-off and handle object, attribute, and relation evaluation in open vocabulary settings.
    \item We evaluate 10 mainstream LVLMs on \data~, focusing on the balance between faithfulness and coverage score. We notice that even GPT-4V(ision) \cite{Achiam2023GPT4TR} still suffers from hallucination, achieving a relatively low faithfulness score despite covering more information within an image compared to other models. 
\end{itemize}

\vspace{-4mm}
\section{Existing LVLMs Hallucination Evaluation Benchmarks and Metrics}
\vspace{-2mm}

As shown in Table~\ref{tab:evaluation_frameworks}, existing studies \cite{li-etal-2023-evaluating, Wang2023EvaluationAA,Zhai2023HallESwitchRA, Lovenia2023NegativeOP,Villa2023BehindTM,Petryk2024ALOHaAN,Kaul2024THRONEAO} have primarily focused on \textit{object-level} hallucination, with only a few recent studies~\cite{faithscore, wang2023llm,Jiang2024HalEvalAU,Zhang2024QuantityMT} recognizing the importance of extending hallucinations to other dimensions. Our benchmark \data~ covers hallucination evaluations of \textit{objects}, \textit{attributes}, and \textit{relations}, and we further detail attributes to color and counting, and relations to positional and comparative, to provide a comprehensive and fine-grained evaluation benchmark. 

Regarding benchmark annotations, many existing benchmarks employ different ways of annotating the evaluation datasets \textit{automatically}. For example, \citet{li-etal-2023-evaluating} employ object detectors to identify all objects in an image; \citet{Zhai2023HallESwitchRA} employ GPT-4V(ision) to generate ground-truth annotations. There are also approaches to developing models specifically for automatic evaluation, thereby bypassing the need for benchmark collections process~\cite{Wang2023EvaluationAA, Gunjal2023DetectingAP}. Given the challenges and potential inaccuracies associated with automated models, our study opts to annotate the evaluation dataset \textit{manually} to ensure the annotation accuracy and encompass the distinct categories of hallucinations. 

Additionally, most existing benchmarks focus exclusively on hallucination evaluation, which can favor precise but uninformative model outputs, overlooking the aspect of coverage. To address the issue, we incorporate coverage scores in our evaluation. We note that two relevant concurrent works~\cite{wang2023llm,Zhai2023HallESwitchRA} also include the coverage scores. However, compared with our work, they are either limited in scope, focusing only on objects or simple attributes and relations, or are unable to be adopted in open-vocabulary generation settings. Besides, along with the benchmark, we propose an evaluation metric generalizing their adopted CHAIR metric.
\section{\data~}
\label{sec:benchmark}

\begin{figure}[t]
 \centering
 \includegraphics[width=\linewidth]{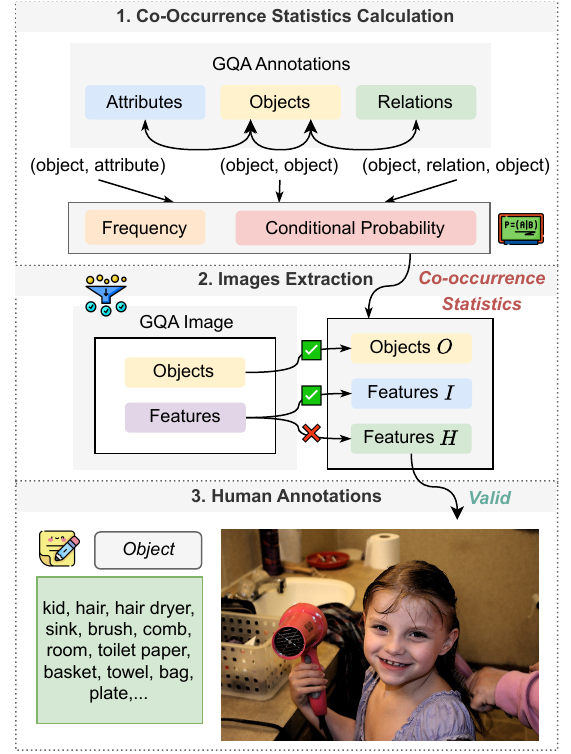}
 \vspace{-7mm}
 \caption{Overview of our proposed benchmark \data~ collection procedure: (1) Image collection (\Cref{sec:images_collection}): (a) \textit{Co-occurrence statistics calculation} (\Cref{sec:dependecies_calculation}): We employ two statistical measures to determine co-occurring features – frequencies and conditional probabilities; (b) \textit{Image extraction} (\cref{sec:extraction_steps}): Next, we leverage the identified co-occurrence statistics to systematically extract images from existing datasets; (2) \textit{Human Annotations} (\Cref{sec:annotation}): Finally, we manually annotate each image within the distinct feature subsets, adhering to the definition in \Cref{sec:definition}. Here, we provide an example of how we use the co-occurrence statistics to select images for \textit{object} subsets and add human annotations for later evaluation.}
 \vspace{-5mm}
 \label{fig:benchmark_overview}
\end{figure}

In this section, we detail the methodology employed to create the benchmark, which aims to evaluate the hallucination issues of LVLMs. As illustrated in Figure~\ref{fig:benchmark_overview}, constructing this benchmark involves two principle phases: the \textit{collection} of images (\Cref{sec:images_collection}) and their subsequent \textit{annotation} (\Cref{sec:annotation}). 

\vspace{-2mm}
\subsection{Image Collection}
\label{sec:images_collection}
\vspace{-1mm}

We aim to select images that can effectively expose the issue of model hallucinations. We hypothesize that when models are repeatedly exposed to specific combinations of features -- such as object existence, object attributes, and object relations -- during training, they develop a pronounced \textit{associative bias}, which leads the models to expect these co-occurring features in similar situations. Consequently, when a model encounters an image containing only one element of a familiar combination, it may erroneously infer the presence of the associated feature. This associative bias is one primary source of model hallucinations \cite{li-etal-2023-evaluating,zhou2023analyzing}. To explore this phenomenon, we initially analyze the co-occurrence statistics of \textit{object-object}, \textit{object-attribute}, and \textit{object-relation-object} combinations within the extensively annotated GQA \cite{Hudson2019GQAAN} dataset. We then curate a collection of images representing \textit{frequently} and \textit{infrequently} co-occur (object, object), (object, attribute), (object, relation, object) tuples. By doing so, we identify the most challenging images to construct a benchmark, to which we then add detailed human annotations for later thorough evaluation. 

We first outline the definition (\Cref{sec:definition}), then explain the process for calculating co-occurrence statistics (\Cref{sec:dependecies_calculation}), and finally describe the steps for using these dependencies to select images (\Cref{sec:extraction_steps}).

\vspace{-1mm}
\subsubsection{Definition}
\label{sec:definition}

We first define three principal features to assess hallucination issues in LVLMs. The first feature, \textbf{Object existence} (object-object), encompasses all visual entities within an image, covering both \textit{foreground} and \textit{background} elements. The second feature, \textbf{Attribute} (object-attribute), focuses on the characteristics of objects, with a particular emphasis on \textit{color} and \textit{counting}. Our analysis within this category is divided into two segments: \textit{object} and \textit{people}. For objects, we concentrate on the color and count of each item not related to people (\textit{e.g.}, six \textit{green apples} on the table). For people, we highlight the colors of attire and the total number of individuals depicted (\textit{e.g.}, a woman who is wearing a \textit{red jacket}). The third feature, \textbf{Relation} (object-relation-object), pertains to the relational information between the objects in the image. Here, we focus on \textit{positional} and \textit{comparative} relation. Specifically, the positional relation tests the relative position between the objects, while the comparative relation analyzes the understanding of ``which object is larger than the other.'' 

\subsubsection{Quantifying Co-Occurring Features}
\label{sec:dependecies_calculation}

To utilize co-occurring features effectively, the first step involves computing the \textit{statistical dependencies} between different features. This analysis aids in identifying dominant co-occurrence patterns in the data, thereby spotlighting features with strong associations that the model might have internalized. We employ two statistical methods to determine these dependencies -- \textit{frequencies} and \textit{conditional probabilities}. \textbf{Frequency} provides insights by quantifying the frequency of specific features in conjunction with particular objects, attributes, or relations, thereby illuminating the raw distribution of these features throughout the dataset. To delve deeper, we calculate the \textbf{conditional probability}, which quantifies the likelihood of encountering a specific feature given the presence of an object:

\vspace{-5mm}
\begin{equation}
    \small
    \mathcal{P}(\text{feature}|\text{object}) = \frac{\text{Frequency}(\text{feature, object)}}{\text{Frequency}(\text{object})},
\end{equation}
where feature $\in$ \{object, attribute, relation\}. Our goal is to identify objects whose conditional probability distributions exhibit significant skew. To achieve this, we explore five distinct metrics based on conditional probabilities. Detailed definitions of these five metrics are provided in Appendix \ref{apx:cond_prob}.

\subsubsection{Utilizing Co-Occurrence Statistics for Image Extraction}
\label{sec:extraction_steps}

Leveraging the identified co-occurrence statistics, we systematically extract images from existing datasets. The process includes several critical steps:

\begin{enumerate}
    \item Identify objects ($\mathbf{O}$) that exhibit the \textit{most pronounced} co-occurrence dependencies, including frequency and conditional probabilities:
    \begin{equation}
        \small
        \mathbf{O} = \{\arg \max_{o} \mathcal{P}(f|o) | f \in \mathcal{F}\},
    \end{equation}
    where $\mathcal{F}$ denotes the set of all features (including object, attribute, and relation) annotated in the dataset, $o$ represents any object annotated in the dataset, and $\mathcal{P}$ signifies all statistical dependencies, including frequencies and five kinds of conditional probabilities.
    
    \item Select features that are \textit{minimally} associated with each identified object in $\mathbf{O}$, denoted as set $\mathbf{I}$,  thereby spotlighting instances where common co-occurrences are \textit{absent}:
    \begin{equation}
        \small
        \mathbf{I} = \{\arg \min_{i} \mathcal{P}(i|o) | i \in \mathcal{F}_o, o \in \mathbf{O} \},
    \end{equation}
    where $\mathcal{F}_o$ denotes the set of all features (including object, attribute, and relation) annotated in the dataset related to object $o$ and $\mathcal{P}$ signifies all statistical dependencies.
    
    \item Determine features that are \textit{most frequently} co-occurring with each identified object in $O$, denoted as set $\mathbf{H}$, serving as \textit{strong} associative tendencies:
    \begin{equation}
        \small
        \mathbf{H} = \{\arg \max_{h} \mathcal{P}(h|o) | h \in \mathcal{F}_o, o \in \mathbf{O} \},
    \end{equation}
    where $\mathcal{F}_o$ denotes the set of all features (including object, attribute, and relation) annotated in the dataset related to object $o$ and $\mathcal{P}$ signifies all statistical dependencies.
    
    \item Collect images $\mathbf{C}$ for each feature in $\mathbf{I}$ corresponding to an object in $\mathbf{O}$, with the chosen images \textit{including} the specified feature and object, yet \textit{excluding} any features from $\mathbf{H}$, to create clear cases for testing the model's associative bias:
    \begin{equation}
        \small
        \mathbf{C} = \{c: (o,f) | o \in \mathbf{O}, f \in \mathbf{I}, \text{and } f \not\in \mathbf{H}\}
    \end{equation}
    where $c$ denotes an image that contains the object $o$ characterized by the feature $f$.
\end{enumerate}

For each feature defined in \Cref{sec:definition}, we adhere to the outlined steps to extract images from the GQA dataset. Subsequently, we manually review the collected images by two expert annotators to ensure that only those of high quality and with clear annotations are retained. These procedures enable us to amass a collection of images for evaluating the object existence and the relations. However, extracting images that accurately represent specific \textit{attributes} proved to be challenging due to the limited attribute annotations in GQA. To overcome this, we source copyright-free images from the Internet\footnote{We use Pixel, a free stock photos platform: \url{https://www.pexels.com/} for image retrieval.}, guided by the attribute-related statistics gathered in the previous step. The statistics of our proposed benchmark are detailed in \Cref{tab:benchmark_stats}.

\vspace{-1mm}
\subsection{Annotation}
\label{sec:annotation}

\begin{figure*}[ht!]
 \centering
 \includegraphics[width=\textwidth]{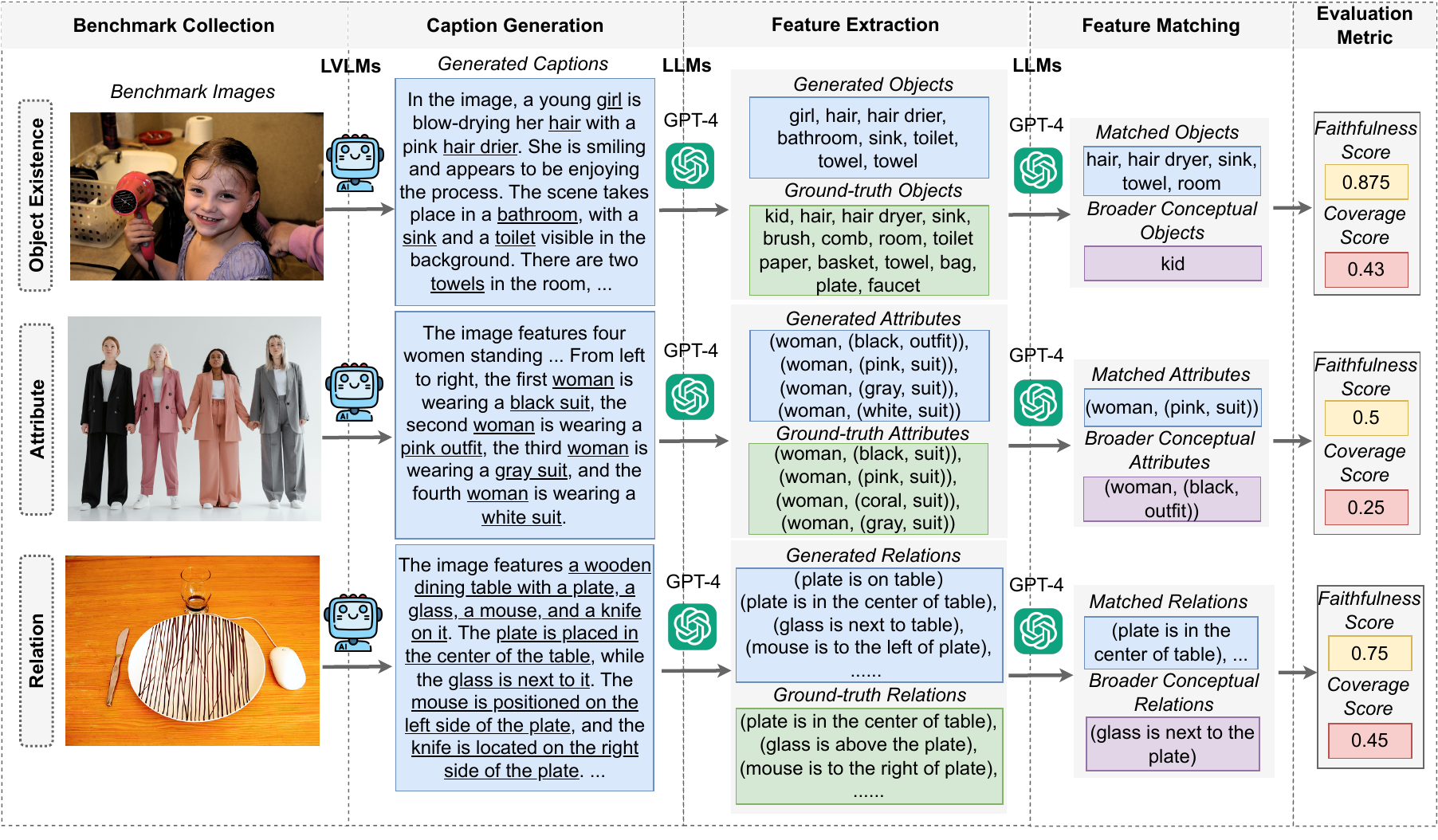}
 \vspace{-7mm}
 \caption{Overview of \eval~ evaluation framework: (1) Firstly, LVLMs generate captions from \data~ benchmark images. (2) Following this, LLMs are employed to \textit{extract} pivotal features that encapsulate from the generated descriptions. (3) Subsequently, these features are \textit{aligned} with a pre-defined list of ground-truth features using LLMs, facilitating the creation of two essential outputs: a dictionary of matched features and a more extensive dictionary encompassing broader conceptual matches. (4) Finally, we calculate two key metrics: \textit{faithfulness} and \textit{coverage}. These metrics measure the LVLMs' comprehension by evaluating how well the generated captions encapsulate the salient features of the images and the breadth of concepts they cover, respectively.}
 \vspace{-2mm}
 \label{fig:evaluation_overview}
\end{figure*}

For each image within the distinct feature subsets, we manually annotate them based on existing annotations, adhering to the definitions discussed in Section \ref{sec:definition}. Figure \ref{fig:benchmark_overview} presents an example in the object subset, while Figure \ref{fig:evaluation_overview} illustrates three examples in the object, attribute, and relation subsets from our collected benchmark. Below, we discuss the details of these annotations.

\paragraph{Object Existence.} Through manual verification of existing annotations, we enhance the dataset by including additional annotations to ensure all visual entities within an image are accounted for. This contains both \textit{foreground} and \textit{background} entities. For example, in an image showing ``a lady sitting on a bench in front of a building,'' the objects to be annotated are the ``lady,'' ``bench,'' and ``building.''

\paragraph{Attributes.} In a similar vein to the approach adopted in the object subset, we further enhance images by appending detailed attribute annotations to the depicted objects. Our analysis within this category bifurcates into two subsets: \textit{object} and \textit{people}. Within the object sub-category, for an image described as ``two green apples on a white table,'' the identified attributes are ``(green, apple)'' for each apple and ``(white, table)'' for the table. For \textit{people} sub-category, in a scene showing ``a woman wearing a red jacket with black shoes,'' the identified attribute is ``(woman, (red, jacket), (black, shoes))''.

\paragraph{Relations.} In our benchmark, we capture \textit{positional} relations between objects. For instance, the statement ``the bed is to the left of the table'' illustrates the positional relation between ``bed'' and ``table''. Conversely, the inverse statement ``the table is to the right of the bed'' is equally valid and is annotated accordingly. Additionally, we annotate descriptions such as ``a bed is on the left side of the image'' to denote the positional relations of objects at the image level. For \textit{comparative} relations, we use an annotation scheme that assigns a numerical rank based on object size, ordering objects from largest to smallest (\textit{e.g.}, ``1. bed, 2. table, 3. cup'').

\begin{table}[t]
\small
\centering
\scalebox{0.97}{
\begin{tabular}{lccc}
\toprule
\textbf{Category} & \textbf{Sub-Category} & \textbf{\# Images} & \textbf{Source}\\
\midrule

Object Existence & - & 50 & GQA\\
\midrule
\multirow{2}{*}{Attribute} & Object & 27 & Pixel\\
& People & 34 & Pixel\\
\midrule
\multirow{2}{*}{Relation} & Positional & 50 & GQA\\
& Comparative & 50 & GQA\\

\bottomrule
\end{tabular}
}
\vspace{-2mm}
\caption{In the \data~ benchmark, we categorize images into three main areas: object existence, attributes, and relations, as outlined in \Cref{sec:definition} and \Cref{sec:extraction_steps}. Attributes are further split into \textit{object} (focusing on color and count of each item not related to people) and \textit{people} (emphasizing the attire colors and the total number of individuals. For relations, we examine both \textit{positional} relations between objects and \textit{comparative} sizes. 
}
\label{tab:benchmark_stats}
\vspace{-5mm}
\end{table}

Ultimately, \data~ provides a set of tuples $(I, F_G, p_G)$, where $I$ denotes the image, $F_G$ is the feature annotations of the image, and $p_G$ represents the prompt designed for LVLMs generation. The designed prompts $p_G$ are shown in  \Cref{apx:caption_generation_prompts} for each subset -- object, attribute, and relation.

\begin{table*}[t]
    \small
    \centering
    \renewcommand{\arraystretch}{1.1}
    \setlength{\tabcolsep}{0.6pt}
    \begin{tabular}{l c c c c c c c c c c c c}
    \toprule    
    \multirow{4}{*}{\textbf{Model}}&\multicolumn{2}{c}{\textbf{Object}}&\multicolumn{4}{c}{\textbf{Attribute}}&\multicolumn{4}{c}{\textbf{Relation}}&\multirow{4}{*}{\makecell{Average \\ Faithful. \\Score \\ (\%)}}&\multirow{4}{*}{\makecell{Average \\ Cover. \\Score \\ (\%)}}\\
    \cmidrule(lr){2-3} 
    \cmidrule(lr){4-7}
    \cmidrule(lr){8-11}
    & \multicolumn{2}{c}{\textit{Existence}} & \multicolumn{4}{c}{\textit{Color \& Counting}} & \multicolumn{2}{c}{\textit{Positional}} & \multicolumn{2}{c}{\textit{Comparative}} \\
    \cmidrule(lr){4-7}

& \multicolumn{2}{c}{} & \multicolumn{2}{c}{Object} & \multicolumn{2}{c}{People} & \multicolumn{2}{c}{} & \multicolumn{2}{c}{} \\
    \cmidrule(lr){2-3} 
    \cmidrule(lr){4-5}
     \cmidrule(lr){6-7}
     \cmidrule(lr){8-9}
    \cmidrule(lr){10-11}
&Faithful$_{\uparrow}$&Cover$_{\uparrow}$&Faithful$_{\uparrow}$&Cover$_{\uparrow}$&Faithful$_{\uparrow}$&Cover$_{\uparrow}$&Faithful$_{\uparrow}$&Cover$_{\uparrow}$&Faithful$_{\uparrow}$&Cover$_{\uparrow}$&\\
    \midrule
    InstructBLIP & 74.5 & 24.8 & 72.0 & 23.9 & 47.1 & 9.3 & 50.0 & 13.6 & 66.9 & 35.6 & 62.1 & 21.44 \\
    LLaVA-1.5 & 72.1  & 24.7 & 74.6 & 37.8 & 43.3 & 12.1 & 64.8 & 14.9 & 51.9 & \hlc{lightblue}{40.1~} & 61.34 & 25.92 \\ 
    MiniGPT-4 v2  & 65.0 & 25.4 & \hlc{lightyellow}{64.5~} & 17.9 & 38.9 & 11.6 & \hlc{lightyellow}{38.8~} & \hlc{lightblue}{33.1~} & 44.7 & \hlc{lightyellow}{11.2~} & \hlc{lightyellow}{50.38~} & 19.84 \\
    mPLUG-Owl2 & 71.5 & 24.8 & \hlc{lightblue}{79.9~} & 32.7 & 39.7 & 16.2 & 45.2 & 10.8 & \hlc{lightyellow}{41.6~} & 30.6 & 55.58 & 23.02 \\
    BLIVA & 77.7 & 21.9 & 73.3 & 24.3 & 37.6 & 11.6 & 39.5 & 9.7 & 68.0 & 29.9 & 59.22 & 19.48 \\
    CogVLM  & 71.2 & 35.5 & 75.3 & 24.3  & 43.7 & 22.4 & 51.9 & 10.5 & 49.0 & 35.9 & 58.22 & 25.72 \\
    InternLM-XComposer2 & 82.5 & 23.9  & 75.8 & 26.3 & 50.4 & 13.8 & 62.6 & 11.1 & 64.1 & 38.4 & 67.08 & 22.7 \\
    Qwen-VL-Chat & 70.6 & 28.4 & 75.1 & \hlc{lightblue}{38.6~} & 38.8 & 16.0 & 56.9 & 8.5 & 51.9 & 24.3 & 58.66 & 23.16 \\ 
    Emu2 & \hlc{lightblue}{94.2~} & \hlc{lightyellow}{14.1~} & 66.7 & \hlc{lightyellow}{10.4~} & \hlc{lightblue}{54.3~} & \hlc{lightyellow}{1.9~} & \hlc{lightblue}{72.2~} & \hlc{lightyellow}{1.8~} & \hlc{lightblue}{87.5~} & 12.3 & \hlc{lightblue}{74.98~} & \hlc{lightyellow}{8.1~} \\
    GPT-4V & \hlc{lightyellow}{61.6} & \hlc{lightblue}{38.8~} & 78.5 & 36.3 & \hlc{lightyellow}{34.7~} & \hlc{lightblue}{23.8~} & 46.7 & 12.6 & ~~51.6$^*$ & ~~28.5$^*$ &54.62 & \hlc{lightblue}{28.0~} \\
    \bottomrule
    \end{tabular}
    \vspace{-2mm}
    \caption{The overall evaluation results of object existence, attribute, and relation hallucination in \data~ using GPT-4 as the LLM Agent within \eval~. The highest is highlighted in \hlc{lightblue}{blue}, while the worst performance is highlighted in \hlc{lightyellow}{yellow}. Faithfulness and coverage scores are in percentage (\%). For images that contain people, GPT-4V refrains from generating comments, and we marked this score with an asterisk ($*$).}
    \vspace{-4mm}
    \label{tab:evaluation_results}
\end{table*}

\section{\eval~} 
\label{sec:evaluation_framework}

We propose a framework \eval~ that generalizes CHAIR, a metric that is widely adopted in existing studies~\cite{Zhai2023HallESwitchRA, wang2023llm}, by introducing semantic matching and incorporating both the \textit{faithfulness} and \textit{coverage} aspects into the evaluation. As shown in Figure~\ref{fig:evaluation_overview}, our evaluation process has two steps: \textit{feature extraction} and \textit{matching} (\Cref{sec:features_extract_match}) and \textit{scoring} (\Cref{sec:evaluation_metrics}).

\vspace{-2mm}
\subsection{Feature Extraction and Matching}
\label{sec:features_extract_match}

We start the process by generating an initial response, denoted as $R$, using a specific LVLM with the input pair $(I, p_G)$, where $I$ denotes the image and $p_G$ represents the prompt designed for LVLMs generation from \data~. Then, we leverage an LLM to analyze $R$ and extract key features. This is achieved through a series of prompts $p_E$, outlined in \Cref{apx:features_extraction_prompts}, which are designed to \textit{extract} features from object existence, attributes, and relations, respectively, resulting in a comprehensive list of extracted features from $R$, denoted as $F_R=\{f_{R_1}, f_{R_2}, ..., f_{R_m}\}$. Next, we utilize an LLM to \textit{align} the extracted features list $F_R$ with a pre-annotated ground-truth features list $F_G=\{f_{G_1}, f_{G_2}, ..., f_{G_m}\}$ from \data~. This alignment is facilitated through a set of carefully crafted prompts $p_M$, outlined in \Cref{apx:features_matching_prompts}, tailored to each feature subset, aiming to identify correlations and correspondences. Unlike previous evaluation metrics that rely on a fixed feature list and direct mapping, our approach eschews pre-processing and instead utilizes LLMs' language comprehension capabilities to semantically match extracted features with their ground-truth counterparts. This process yields two key outputs: \textbf{matched features} dictionary ($D_{M}$) and \textbf{broader conceptual matches} dictionary ($D_{B}$).\looseness=-1

$D_{M}$ contains features $f_{R_{i'_m}}$ from $f_R$ that semantically aligned with the features $f_{G_{i_m}}$ from $F_G$, ensuring \textit{precision}. For example, if we have the extracted ``(plaid, shirts)'' and the candidate ground-truth feature is ``(checkered, shirt),'' we can establish a match between these two because ``plaid'' and ``checkered'' are conceptually similar patterns often used interchangeably in the context of textiles. 

$D_{B}$ includes features $f_{R_{j'_n}}$ from $f_R$ that have broader conceptual meanings than the features $f_{G_{j_n}}$ from $F_G$, adding \textit{conceptual depth} to the evaluation. For instance, if we have the extracted ``(red, clothes)'' from an image, and the ground-truth annotation is ``(red, dress),'' we can still consider these features to match. This is because ``clothes'' is a broader category that encompasses ``dress.'' Therefore, despite the slight difference in specificity, the extracted features can be aligned with the ground-truth annotations based on their semantic relationship, where ``dress'' is a sub-type of ``clothes.''

\subsection{Evaluation Metrics}
\label{sec:evaluation_metrics}

We introduce two metrics to evaluate the hallucinations in two dimensions: \textit{faithfulness} and \textit{coverage} based on the original CHAIR metric. 

\paragraph{Faithfulness.} In the context of image captioning, faithfulness measures how closely captions match an image's content, emphasizing \textit{accuracy} in depicting visual elements and their attributes and relations without introducing hallucinations. It is calculated by comparing generated features against actual image features, considering both \textit{direct} ($D_{M}$) and \textit{broader} conceptual similarities ($D_{B}$):
\begin{equation}
    \small
    \text{Faithfulness}(R, F_G) = \frac{|D_{M} \cup set(D_{B})|}{|F_R|} \in [0,1].
\end{equation}

\paragraph{Coverage.} It measures the \textit{comprehensiveness} of the generated captions in capturing the key elements and attributes depicted in the image. It evaluates the proportion of ground-truth features that are successfully captured in the generated response, only through \textit{direct} matches ($D_{M}$):
\begin{equation}
    \small
    \text{Coverage}(R, F_G) = \frac{|set(D_{M})|}{|F_G|} \in [0,1].
\end{equation}
\section{Experiment}

In this section, we perform experiments to evaluate different existing LVLMs within our proposed framework (\Cref{subsec:exp_eval}). We also present evidence demonstrating that our evaluation methodology aligns closely with human judgment (\Cref{subsec:exp_metric}). Additionally, we explore the significance of each design aspect of our framework through ablation studies (\Cref{subsec:exp_ablation}). Finally, we showcase qualitative examples to illustrate our findings (\Cref{subsec:qualitative}).

\subsection{Model Coverage-Faithfulness Evaluation}
\label{subsec:exp_eval}
We use the framework \eval~ to evaluate various LVLMs listed in \Cref{tab:lvlm_models} in the Appendix \ref{apx:lvlms}, employing GPT-4 as the evaluation LLM agent. 

In the evaluation of various models, as shown in \Cref{tab:evaluation_results}, Emu2 distinguishes itself by achieving the highest average faithfulness score of 74.98, signifying its consistent capability to generate responses that accurately reflect the content of the input image. However, Emu2's performance in terms of coverage is less impressive, with the lowest average score of 8.1, suggesting that its responses, while accurate, may not comprehensively cover all elements of the image. When broken down into specific dimensions, Emu2 excels in faithfulness across categories -- scoring 94.2 in object existence, 54.3 in attribute-people, 72.2 in relation-positional, and 87.5 in relation-comparative. Conversely, it lags in coverage, with scores of 14.1 in object existence, 10.4 in attribute-object, 1.9 in attribute-people, and 1.8 in relation-positional. These results point to a potential trade-off between faithfulness and coverage in Emu2's design, where the model prioritizes accuracy at the expense of a broader scope in its responses. This pattern supports the initial hypothesis that \textit{some LVLMs may intentionally sacrifice coverage to improve the precision of their outputs}.

Meanwhile, GPT-4V(ision) distinguishes itself with an unparalleled average coverage score of 28.0, showcasing its adeptness in encapsulating a wide array of features from the input image. This indicates that GPT-4V excels in recognizing and addressing diverse elements within images, although it does not necessarily always maintain the highest accuracy, as seen in its lower faithfulness score of 61.6. Particularly in evaluations concerning the existence of objects, GPT-4V leads with the highest coverage score of 38.8, underlining its comprehensive approach to object detection. This approach tends to favor inclusivity, which might lead to the occasional identification of objects that are not present in the image. Furthermore, in evaluations focused on attributes related to people, GPT-4V again achieves the highest coverage score of 54.3. However, this comes with a trade-off, as it also exhibits a higher tendency towards hallucinations compared to other models, indicating a propensity to generate details or elements that may not be grounded in the actual content of the image.

Models such as LLaVA-1.5 and CogVLM showcase a more equitable performance, achieving respectable scores in both faithfulness and coverage metrics. This highlights their capability to provide responses that are not only precise but also encompassing. Notably, LLaVA-1.5 stands out for its remarkable outcomes, achieved through the efficient use of training data, underscoring the significance of leveraging high-quality instruction-tuning data to enhance model performance.

\vspace{-1mm}
\subsection{Effectiveness of Evaluation Framework}
\label{subsec:exp_metric}
\vspace{-1mm}

\begin{table}[t]
\small
\centering
\setlength{\tabcolsep}{0.8mm}{
\scalebox{0.92}{
\begin{tabular}{lccc}
\toprule
\textbf{Category} & \textbf{Sub-Category} & \textbf{\ Faithful. ($\rho$) } & \textbf{Cover ($\rho$) }\\
\midrule

Object Existence & - & 0.91 & 0.89\\
\midrule
\multirow{2}{*}{Attribute} & Object & 0.99 & 0.98\\
& People & 0.98 & 0.96\\
\midrule
\multirow{2}{*}{Relation} & Positional & 0.78 & 0.86 \\
& Comparative & 0.92 & 0.98 \\

\bottomrule
\end{tabular}
}
}
\vspace{-2mm}
\caption{Pearson correlation ($\rho$) between our GPT-4-based evaluation framework \eval~ and human judgements.}
\vspace{-6mm}
\label{tab:human correlation}
\end{table}

To demonstrate the \textit{effectiveness} and \textit{reliability} of our LLM-based automatic evaluation pipeline, we conduct experiments to evaluate if our evaluation framework correlates with human evaluations in both faithfulness and coverage dimensions. Specifically, we have human and our GPT-4-based evaluation method evaluate InstructBLIP outputs and compute the Pearson correlation ($\rho$) score\footnote{We opt for Pearson correlation as our assessment metric due to its suitability for measuring \textit{linear} relationships, as opposed to Spearman's rank correlation, which is more attuned to \textit{monotonic} relationships.}. As shown in \Cref{tab:human correlation}, for object existence, the findings reveal a significantly strong Pearson correlation of 0.91 for faithfulness and 0.89 for coverage, effectively rejecting the null hypothesis that posits no correlation between the two evaluation methodologies, with a compelling $p$-value of 0. Additionally, our study achieved a notably high correlation of 0.98 in attribute recognition and comparative relations. When evaluating positional relations, which tend to involve longer and more complex descriptions, the correlation scores were not as high as those observed in the other categories but still indicated a very high level of correlation, with 0.78 in faithfulness and 0.86 in coverage. These results affirm the comparability of our automatic evaluation metrics to human evaluation in terms of both \textit{efficacy} and \textit{reliability}.

\subsection{Ablation Study}
\label{subsec:exp_ablation}

In this section, we serve to answer \textbf{two} questions and discuss our findings. 

\begin{table}[t]
\small
\centering
\setlength{\tabcolsep}{0.9mm}{
\scalebox{0.99}{
\begin{tabular}{llll}
\toprule
\textbf{Model} & \textbf{InstructBLIP} & \textbf{LLaVA-1.5} & \textbf{GPT-4V}\\
\midrule

\multicolumn{4}{l}{\textit{Evaluation data: randomly selected}} \\
Faithfulness & 76.5 & 84.5 & 64.1 \\
Coverage  &24.3 & 26.3 & 41.2\\
\midrule 
\multicolumn{4}{l}{\textit{Evaluation data: co-occurrence selected} (\textbf{Ours}) }  \\
Faithfulness &74.5 \textcolor{red}{(-2.0)} & 72.1 \textcolor{red}{(-12.4)} & 61.6 \textcolor{red}{(-2.5)}\\
Coverage  &24.8 \textcolor{red}{(+0.5)} & 24.7 \textcolor{red}{(-1.6)} & 38.8 \textcolor{red}{(-2.4)}\\
\bottomrule
\end{tabular}
}
}
\vspace{-2mm}
\caption{Model performance comparison on our data selection method against random selection. Faithfulness and coverage scores are in percentage (\%).} 
\vspace{-6mm}
\label{tab:co-ocurrence ablate}
\end{table}

1. \textbf{How does our co-occurrence data selection method compare to other alternatives?} 

To illustrate the effectiveness of the co-occurrence data selection method, we set up a baseline of randomly selecting 50 images in the GQA validation split and applying human annotations, the same as for our dataset. For the ablation study, we focus on the well-studied object hallucination. We evaluate three popular models representing query tokens-based image features (InstructBLIP), linear projection-based features (LLaVA-1.5), and advanced commercial LVLMs (GPT-4V). As shown in \Cref{tab:co-ocurrence ablate}, all models tend to produce more hallucinations and exhibit significantly \textit{lower faithfulness} compared to our benchmark. Notably, LLaVA-1.5 scores 12.4 points lower in faithfulness when evaluated against our benchmark. This suggests that our benchmark is challenging due to its reliance on co-occurrence selection. Additionally, the coverage scores for both LLaVA-1.5 and GPT-4V decreased. Upon further analysis through human review, we discover that our benchmark, on average, contains 1.69 more objects than images selected at random. This finding indicates that our data selection method can incorporate more complex objects compared to the random selection approach commonly used in other benchmark constructions.   

2. \textbf{How does our LLM-based evaluation framework compare with LLM-free evaluation?} 

We compare our proposed LLM agent augmented framework against the original CHAIR metric which is adopted by all previous studies. Because the CHAIR metric is limited to evaluating only 80 objects from the MSCOCO dataset, for a fair comparison, we randomly select 20 COCO images and re-annotate them for analysis alongside the CHAIR metric. We have made these annotations publicly available, adhering to the same list of synonyms used in the original CHAIR metric. To conduct this comparison, we utilize two accuracy scores. For Acc (F), we assess the performance by comparing the number of hallucinated objects identified by the metric against the ground-truth hallucinated objects in the caption. If an object is incorrectly identified as hallucinated when it is not, the metric imposes a penalty of -1. This score aligns with the \textit{matching} phase of our framework, ensuring a thorough evaluation of hallucination detection accuracy. For Acc (C), we calculate the number of objects detected by metric over the unique objects mentioned in the caption, assessing our \textit{extraction} phase's efficiency. As shown in Table \ref{tab:LLM agent framework ablate}, our framework significantly outperforms in both faithfulness and coverage accuracy by a large amount. This improvement is due to our framework's \textit{open vocabulary matching} ability, unlike the original CHAIR approach that struggles with new expressions without pre-defined synonyms. Notably, with complex models like GPT-4V, CHAIR's faithfulness accuracy drops to 5.88, highlighting our method's strength in managing diverse object descriptions.

\begin{table}[t]
\small
\centering
\setlength{\tabcolsep}{0.7mm}{
\scalebox{0.94}{
\begin{tabular}{lllll}
\toprule
\textbf{Metric} & \textbf{F.}$_{\uparrow}$ & \textbf{C.}$_{\uparrow}$ & \textbf{Acc (F)}$_{\uparrow}$& \textbf{Acc (C)}$_{\uparrow}$   \\
\midrule
\multicolumn{5}{l}{\textit{Model: InstructBLIP}} \\
CHAIR & 75.0 & 34.3 & 11.11 & 80.66\\
CHAIR$_{\text{LLM}}$ \textbf{(Ours)} & 76.9  & 30.4 & 88.89 \textcolor{red}{(+77.78)} & 100.0 \textcolor{red}{(+19.34)}\\
\midrule
\multicolumn{5}{l}{\textit{Model: LLaVA-1.5}} \\
CHAIR & 74.3 & 34.1 & 30.00 & 83.52\\
CHAIR$_{\text{LLM}}$ \textbf{(Ours)} &81.5 & 27.0 & 90.00 \textcolor{red}{(+60.00)} &   97.08 \textcolor{red}{(+13.56)}\\
\midrule
\multicolumn{5}{l}{\textit{Model: GPT-4V}} \\
CHAIR & 79.3 & 54.8 & 5.88 & 82.35 \\
CHAIR$_{\text{LLM}}$ \textbf{(Ours)} &69.7&57.9 &82.35 \textcolor{red}{(+76.47)}&98.17 \textcolor{red}{(+15.82)}\\
\bottomrule
\end{tabular}
}
}
\vspace{-2mm}
\caption{Comparison of LLM-augmented CHAIR with original CHAIR metric. Here, F. and C. denote faithfulness and coverage scores in percentage (\%). Acc (F) represents the average percentage of hallucinated objects detected by the metric. Acc (C) denotes the average percentage of objects detected by metric.}
\vspace{-5mm}
\label{tab:LLM agent framework ablate}
\end{table}

Moreover, the limitation of CHAIR's pre-defined object list extends to its inability to account for potential hallucinated objects, which are essential for differentiating between mere words and actual objects in captions. This leads to its failure in detecting hallucinated objects, resulting in performance degradation. In contrast, our method overcomes this by using an automatically extracted object list that dynamically matches objects, avoiding this limitation. Although approaches like \citet{wang2023llm} attempt to address this by including a selection of potential hallucinated objects, they cannot guarantee coverage of all possible hallucinated objects, particularly in complex outputs from advanced LVLMs that generate extensive captions.

\subsection{Qualitative Results}
\label{subsec:qualitative}
\vspace{-1mm}

We illustrate the qualitative results of three representative models in \Cref{fig:example_object}, \Cref{fig:example_pos_rela} and \Cref{fig:example_comp_rela} in the Appendix \ref{apx:qualitative_results}. Each model exhibited instances of hallucination in these examples from our evaluation benchmark \data~. Notably, while GPT-4V generates the most comprehensive results, it is also more prone to producing hallucinations. 
\vspace{-2mm}
\section{Conclusion}
\vspace{-2mm}

We introduce a comprehensive multi-dimensional benchmark, named \data~, dedicated to the evaluation of LVLMs, with a particular focus on measuring hallucinations in generative tasks. Our benchmark categorizes hallucinations into three distinct types -- object, attribute, and relation -- offering a detailed understanding of model inaccuracies. Furthermore, our novel evaluation framework, referred to as \eval~, employs a two-stage approach that integrates an LLM, effectively addressing the complexities related to open vocabularies, semantic similarities, and the intricate assessment of attributes and relationships. This method significantly enhances the precision and depth of image captioning evaluations compared to previous methods. Our experimental findings highlight the persistent challenges in this field, demonstrating that even state-of-the-art models such as GPT-4V, are prone to a considerable degree of hallucination. This study emphasizes the imperative for continuous advancements in LVLM evaluation techniques and establishes a new benchmark for future endeavors aimed at reducing hallucination and bolstering the reliability of content generated by LVLMs.
\section{Ethical Considerations}

Our work investigates the phenomenon of hallucinations in outputs generated by LVLMs. Here, we outline the primary ethical considerations associated with our study. In developing our evaluation framework, we employed GPT-4 for feature extraction and matching tasks to evaluate the model's hallucination. Consequently, we recognize that any biases inherent to the GPT-4 model will likely influence the results observed in our benchmark \cite{Achiam2023GPT4TR,huang2023embrace,qiu-etal-2023-gender,Wang2024NewJN}. Furthermore, our data collection efforts encompassed datasets from GQA and images sourced from the internet (specifically Pixel\footnote{\url{https://www.pexels.com/}}). We acknowledge and adhere to the pertinent policies and requirements governing data sharing and utilization within our benchmark. 
\section{Limitations}

Our humanly annotated benchmark, \data~, provides a more comprehensive and detailed evaluation than previous works in objects, attributes, and relations. This dataset is humanly curated to cover a broad spectrum of hallucination phenomena, focusing on object existence, color and count attributes, and positional and comparative relations. Despite the extensive coverage, it is essential to acknowledge that we did not fully address the entire range of possible attributes and relations that could be subject to hallucination in LVLMs. Although not covered in our current benchmark, additional elements are equally crucial for a holistic understanding and assessment of LVLMs. Further, we employ a single prompt for evaluating LVLM performance. This approach raises the possibility that some models may not be adequately trained to follow these instructions as intended or require refined prompt engineering to achieve optimal performance. 
\section{Acknowledgment}

We thank anonymous reviewers for their helpful
feedback. We also thank members from the UCLA NLP group for their
feedback and discussions. This research is supported by Meta Sponsor Research Award, Okawa Foundation Research Grant, Google Research Scholar, Amazon Alexa AI Research Award, and a gift from UCLA Institute for Technology, Law and Policy. This material is based on research supported by the ECOLE program under Cooperative Agreement HR00112390060, both with the US Defense Advanced Research Projects Agency (DARPA).

\bibliography{anthology,custom}
\bibliographystyle{acl_natbib}

\clearpage
\appendix

\begin{table*}[h]
\small
\centering
\scalebox{1.0}{
\begin{tabular}{lccc}
\toprule
 \textbf{Model} & \textbf{Visual Encoder} & \textbf{Alignment Network} & \textbf{Language Model} \\
\midrule
InstructBLIP & $\text{EVA CLIP ViT-G/14}_{\text{1.1B}}$ & Q-Former & $\text{Vicuna}_{\text{7B}}$ \\
LLaVA-1.5 & $\text{CLIP ViT-L/14-336px}_{\text{0.4B}}$ & MLP & $\text{Vicuna-v1.5}_{\text{13B}}$ \\
MiniGPT-v2 &  $\text{EVA CLIP ViT-G/14}_{\text{1.1B}}$ & Linear Projection & $\text{LLaMA-2}_{\text{7B}}$ \\
mPLUG-Owl2 & $\text{CLIP ViT-L/14}_{\text{0.4B}}$ & Cross Attention & $\text{LLaMA-2}_{\text{7B}}$ \\
BLIVA & $\text{EVA CLIP ViT-G/14}_{\text{1.1B}}$ & Q-Former \& Linear Projection & $\text{Vicuna}_{\text{7B}}$ \\
CogVLM & $\text{EVA2-CLIP-E/14}_{\text{4.7B}}$  & MLP &  $\text{Vicuna-v1.5}_{\text{7B}}$ \\
InternLM-Xcomposer2 & $\text{CLIP ViT-L/14-336px}_{\text{0.4B}}$ & Partial Low-Rank Adaptation  &$\text{InternLM2}_{\text{7B}}$  \\
Qwen-VL & $\text{CLIP ViT-G/14}_{\text{1.9B}}$ & Cross Attention & $\text{QwenLM}_{\text{13B}}$ \\
Emu2 & $\text{EVA2-CLIP-E-plus/14}_{\text{5.0B}}$ & Linear Projection &$\text{LLaMA}_{\text{33B}}$  \\
GPT-4(V) & Unknown & Unknown & GPT-4 \\

\bottomrule
\end{tabular}
}
\caption{Architectures of mainstream LVLMs evaluated in our benchmark. InstructBLIP \cite{instructblip}, LLaVA-1.5 \cite{liu2023improvedllava}, MiniGPT-v2 \cite{Chen2023MiniGPTv2LL}, mPLUG-Owl2 \cite{Ye2023mPLUGOwl2RM}, BLIVA \cite{hu2023bliva}, CogVLM \cite{wang2024cogvlm}, InternLM-XComposer2 \cite{dong2024internlmxcomposer2}, Qwen-VL \cite{Bai2023QwenVLAV}, Emu2 \cite{sun2023generative} and GPT-4V \cite{Achiam2023GPT4TR}.} \label{tab:lvlm_models}
\end{table*}

\section{Large Vision-Language Models}
\label{apx:lvlms}

The recent advancements in large language models (LLMs) \cite{Achiam2023GPT4TR,Touvron2023LLaMAOA, Touvron2023Llama2O,Vicuna,Bai2023QwenVLAV} have sparked a wave of research focused on enhancing vision-language pre-trained models (VLPMs) \cite{Kim2021ViLTVT, Alayrac2022FlamingoAV, Li2023BLIP2BL}. 
By incorporating the versatile capabilities of LLMs, these studies aim to improve the language understanding and generation abilities of VLPMs significantly. In this paper, we refer to the enhanced VLPMs with the integration of LLMs as \textit{Large Vision-Language Models} (LVLMs) \cite{li-etal-2023-evaluating}. LVLMs excel in comprehending both the visual semantics of objects in images and the linguistic semantics associated with these objects by leveraging the extensive parametric knowledge embedded in the LLMs. This dual understanding enables LVLMs to conduct intricate reasoning about the concepts related to these objects. Consequently, LVLMs demonstrate strong performance in various traditional multi-modal tasks, such as visual question answering, image captioning, and object detection, highlighting their versatility and robustness in these domains \cite{liu2023llava, zhu2023minigpt, ye2023mplugowl,instructblip, liu2023improvedllava, hu2023bliva, Achiam2023GPT4TR, huang2023lvlms, huang2024pixels}. \Cref{tab:lvlm_models} shows comparison of these LVLMs.

\section{Conditional Probabilities}
\label{apx:cond_prob}

\begin{enumerate}
    \item $\mathcal{P}(\text{feature}|\text{object})_\text{max}$: maximum conditional probability, highlighting the strongest feature-object associations.
    \item $\mathcal{P}(\text{feature}|\text{object})_\text{avg}$: average conditional probability, offering a broad view of how features tend to cluster around objects.
    \item $\mathcal{P}(\text{feature}|\text{object})_\text{max} -\mathcal{P}(\text{feature}|\text{object})_\text{avg}$: the difference between the maximum and average conditional probabilities, revealing objects with outlier features.
    \item $\mathcal{P}(\text{feature}|\text{object})_\text{avg} -\mathcal{P}(\text{feature}|\text{object})_\text{min}$: the spread between average and minimum conditional probabilities, indicating the range of commonality among features.
    \item $\mathcal{P}(\text{feature}|\text{object})_\text{max} -\mathcal{P}(\text{feature}|\text{object})_\text{min}$: the range between maximum and minimum conditional probabilities, capturing the full spectrum of feature variability.
\end{enumerate}

\section{Captions Generation Prompts}
\label{apx:caption_generation_prompts}

\begin{itemize}

    \item Object Existence: Write a detailed description of the image. Provide information about all objects in front and background.
    \item Attribute (Object): Write a detailed description of the image. Provide information about the total number and colors of all objects from left to right and up to bottom.
    \item Attribute (People): Write a detailed description of the image. Provide information about the total number of people and colors of clothes for each person from left to right.
    \item Relation (Positional): Describe the positional relationship between all the objects in the image in detail, using left, right, top, and bottom etc, from the view of the observer.
    \item Relation (Comparative): Rank the size of all the objects in the image in detail, from large to small.
\end{itemize}

\section{Features Extraction Prompts}
\label{apx:features_extraction_prompts}

The feature extraction prompts for objects, color and counting attributes, positional relation and comparative relation are illustrated in \Cref{tab:extraction_obj}, \Cref{tab:extraction_att_obj}, \Cref{tab:extraction_att_peo},  \Cref{tab:extraction_pos}, and  \Cref{tab:extraction_comparative}, respectively. 

\begin{table*}[h!]\centering
\begin{minipage}{0.95\textwidth}   
\centering
\begin{tcolorbox} 
    \centering
    \small
    \begin{tabular}{p{0.95\textwidth}}
   { {\bf System message} } \\
    You are a language assistant who helps extract information from given sentences. \\
    \midrule
    {\bf Prompt}  \\
    Given an image with a caption that is generated by a vision-language model. 
    
    Please act as a linguistic master and extract all the objects from the captions. 
    
    Format your response in JSON format, with the key being ``objects'' and the value being a list of objects. 
        
    Please only extract objects without including attributes. For example, extract ``field'' instead of ``grassy field''. Also be mindful of plural forms. For example, extract "cow" instead of ``cows''.
    
    Please only extract the object that is a concrete entity in the real world instead of abstract concepts, actions, and moves.
    
    It cannot be an abstract notion such as day, time, scene, moment, image, game, sport, setting, plot, atmosphere, surroundings, group etc.
    
    It cannot be any words describing the emotions such as excitement, enthusiasm, etc.
    
    It cannot be any words describing the positions in the image, such as foreground, background, left, right, etc. \\\\
    
    For clarity, consider these examples: \textcolor[rgb]{0,0.7,0}{ \{In-context examples\} } \\
------------------- \\
    With these examples in mind, please help me extract the objects based on the factual information in the caption. \\
    Here is the caption: 
\textcolor[rgb]{0.8,0,0}{\{Input Caption\}} \\
\\

    \end{tabular}
\end{tcolorbox}
\caption{Prompt template for extracting \textbf{objects}. \textcolor[rgb]{0,0.7,0}{ \{In-context examples\} } are in-context examples. \textcolor[rgb]{0.8,0,0}{\{Input caption\}} are captions generated by evaluated models.}
    \label{tab:extraction_obj}
\end{minipage}
\end{table*}

\begin{table*}[h!]\centering
\begin{minipage}{0.95\textwidth}   
\centering
\begin{tcolorbox} 
    \centering
    \small
    \begin{tabular}{p{0.95\textwidth}}
   { {\bf System message} } \\
    You are a language assistant who helps extract information from given sentences. \\
    \midrule
    {\bf Prompt}  \\
    Given an image with a caption that is generated by a vision-language model.
    
    Please act as a linguistic master and extract the total number and colors of all objects as mentioned in the captions.
    
    Your answer should be a dictionary of this format: \{``total num of objects'': ``(NUM, OBJECT)'', ``objects'': \{``ORDER'': ``(ATTRIBUTE, OBJECT)''\}\}. Remember OBJECT should be in singular format. \\\\

    For clarity, consider these examples: \textcolor[rgb]{0,0.7,0}{ \{In-context examples\} } \\
------------------- \\
        With these examples in mind, please help me extract the objects and attributes based on the factual information in the caption. \\
        Here is the caption: 
\textcolor[rgb]{0.8,0,0}{\{Input Caption\}} \\
\\

    \end{tabular}
\end{tcolorbox}
\caption{Prompt template for extracting \textbf{attributes (object)}. \textcolor[rgb]{0,0.7,0}{ \{In-context examples\} } are in-context examples. \textcolor[rgb]{0.8,0,0}{\{Input caption\}} are captions generated by evaluated models.}
    \label{tab:extraction_att_obj}
\end{minipage}
\end{table*}

\begin{table*}[h!]\centering
\begin{minipage}{0.95\textwidth}  
\centering
\begin{tcolorbox} 
    \centering
    \small
    \begin{tabular}{p{0.95\textwidth}}
   { {\bf System message} } \\
    You are a language assistant who helps extract information from given sentences. \\
    \midrule
    {\bf Prompt}  \\
    Given an image with a caption that is generated by a vision-language model.
    
    Please act as a linguistic master and extract the total number of people and colors of clothes for each person as mentioned in the captions.
    
    Your answer should be a dictionary of this format: \{``total num of people'': ``(NUM, PERSON)'', ``clothes'': \{``ORDER'': ``person'': ``PERSON'', ``object'': ``(ATTRIBUTE, OBJECT)'', ``action'': ``ACTION''\}\}. OBJECT can be clothes or accessories (e.g., bags, socks). \\\\

    For clarity, consider these examples: \textcolor[rgb]{0,0.7,0}{ \{In-context examples\} } \\
------------------- \\
        With these examples in mind, please help me extract the objects and attributes based on the factual information in the caption. \\
        Here is the caption: 
\textcolor[rgb]{0.8,0,0}{\{Input Caption\}} \\
\\

    \end{tabular}
\end{tcolorbox}
\caption{Prompt template for extracting \textbf{attributes (people)}. \textcolor[rgb]{0,0.7,0}{ \{In-context examples\} } are in-context examples. \textcolor[rgb]{0.8,0,0}{\{Input caption\}} are captions generated by evaluated models.}
    \label{tab:extraction_att_peo}
\end{minipage}
\end{table*}

\begin{table*}[h!]\centering
\begin{minipage}{0.95\textwidth}  
\centering
\begin{tcolorbox} 
    \centering
    \small
    \begin{tabular}{p{0.95\textwidth}}
   { {\bf System message} } \\
You are a language assistant that helps to extract information from given sentences. \\
    \midrule
    {\bf Prompt}  \\
Given an image with a caption that is generated by a vision language model. \\
        Please act as a linguistic master and extract a set of words describing the spatial or positional relations between all the visual objects from the captions.
        Your answer should be a list of values that are in format of object1 relation with object2 with the relation being left, right, top, bottom, middle etc.
        Do not extract the attribute along with the object and don't extract any relation that is an verb, replace it with simply which object is (on or to the left or etc) the other object or the image. Formulate your response into a JSON object with the key being ``relations'' and the value being a list of relations. 
        If there are no relations found, please return an empty list. \\\\
        For clarity, consider these examples: \textcolor[rgb]{0,0.7,0}{ \{In-context examples\} } \\
------------------- \\
With these examples in mind, please help me extract the relations based on the information in the caption.\\
        Here is the caption: 
\textcolor[rgb]{0.8,0,0}{\{Input Caption\}} \\
\\

    \end{tabular}
\end{tcolorbox}
\caption{Prompt template for extracting \textbf{positional relations}. \textcolor[rgb]{0,0.7,0}{ \{In-context examples\} } are in-context examples. \textcolor[rgb]{0.8,0,0}{\{Input caption\}} are captions generated by evaluated models.}
    \label{tab:extraction_pos}
\end{minipage}
\end{table*}

\begin{table*}[h!]\centering
\begin{minipage}{0.95\textwidth}  
\centering
\begin{tcolorbox} 
    \centering
    \small
    \begin{tabular}{p{0.95\textwidth}}
   { {\bf System message} } \\
You are a language assistant that helps to extract ranking from given sentences. \\
    \midrule
    {\bf Prompt}  \\
Given an image with a caption that is generated by a vision language model. \\
             Given an image with a caption that is generated by a vision language model. 
        Please act as a linguistic master and extract the rank of all the objects from large to small as mentioned in the captions.
        Your answer should be a dict of values which the keys represent the ranks starting from 1 and values are the No.1 largest object to smallest.
        If the caption does not mention the order of the object, you can by default view the order of objects appearance as from largest to smallest.
        If there are no objects mentioned in the caption, you can return an empty dict. \\\\
        For clarity, consider these examples: \textcolor[rgb]{0,0.7,0}{ \{In-context examples\} } \\
------------------- \\
With these examples in mind, please help me extract the relations based on the information in the caption.\\
        Here is the caption: 
\textcolor[rgb]{0.8,0,0}{\{Input Caption\}} \\
\\

    \end{tabular}
\end{tcolorbox}
\caption{Prompt template for extracting \textbf{comparative relations}. \textcolor[rgb]{0,0.7,0}{ \{In-context examples\} } are in-context examples. \textcolor[rgb]{0.8,0,0}{\{Input caption\}} are captions generated by evaluated models.}
    \label{tab:extraction_comparative}
\end{minipage}
\end{table*}

\section{Features Matching Prompts}
\label{apx:features_matching_prompts}

The features matching prompts for objects, color and counting attributes, positional relation and comparative relation are illustrated in \Cref{tab:matching_obj}, \Cref{tab:matching_att_obj}, \Cref{tab:matching_att_people}, \Cref{tab:matching_pos}, and  \Cref{tab:matching_comparative}, respectively.

\begin{table*}[h!]\centering
\begin{minipage}{0.95\textwidth} 
\centering
\begin{tcolorbox} 
    \centering
      \small
    \begin{tabular}{p{0.95\textwidth}}
   { {\bf System message} } \\
You are given a task to match objects from two lists that have the same meaning. \\
    \midrule
    {\bf Prompt}  \\
        
        Input Lists:\\
        
        1. ``gt-objects'': Ground truth objects in the image.\\
        2. ``generated-objects'': Objects identified by a vision-language model. \\\\

        Matching Criteria: \\
        - For each object in ``generated-objects'', find the object in the ``gt-objects'' that have the same meaning and add it to the ``matched-objects'' dictionary.
        
        - By the same meaning, we mean the words can be synonyms, can be plural/singular forms of each other and can also have different length of words to express the same meaning of objects, etc.
        
        - Note since we find the matched object for each object in ``generated-objects'', it's ok that multiple objects in``generated-objects'' match one object in ``gt-object'', list all matches.
        
        - There is special scenario that when you can't find the matched object in ``gt-objects'' but you can find one or more object is a subset or a sub category of the generated object, which means that the generated object is a broader concept of the object in ``gt-objects'', add it to the ``broader-concept'' dictionary instead of the ``matched-objects''. If there are many objects are a subset or a sub category of the generated object, you can pick anyone of them. Note we are matching for each object in ``generated-objects''. If you can find the matched object in ``gt-objects'', you should not add it to the ``broader-concept'' dictionary. \\\\
        
        Output:\\
        1.  A ``broader-concept'' dictionary: only if an object from ``generated-objects'' denotes a broader category of a concept in ``gt-objects''. Key = word from ``generated-objects'', Value = word from ``gt-objects''. \\
        2. A ``matched-objects'' dictionary: Key = word from ``generated-objects'', Value = word from ``gt-objects''. 
        It should not contain any words from the ``broader-concept'' dictionary.\\\\
        For clarity, consider these examples: \textcolor[rgb]{0,0.7,0}{ \{In-context examples\} } \\
------------------- \\
        With these examples in mind, please help me extract the broader-concept, and matched-objects from the following two objects lists. \\ 
        1. gt-objects: \textcolor[rgb]{0.8,0,0}{\{Input Ground Truth Objects\}}  \\
        2. generated-objects: \textcolor[rgb]{0.8,0,0}{\{Input Generated Objects\}} \\\\
\\

    \end{tabular}
\end{tcolorbox}
\caption{Prompt template for matching \textbf{objects} in image caption and reference caption.
\textcolor[rgb]{0,0.7,0}{ \{In-context examples\} } are in-context examples. \textcolor[rgb]{0.8,0,0}{\{Input Ground Truth Objects\}} are the ground truth objects list
\textcolor[rgb]{0.8,0,0}{\{Input Generated Objects\}} are the extracted objects list from the extraction step which are originally captions generated by evaluated models.}
    \label{tab:matching_obj}
\end{minipage}
\end{table*}

\begin{table*}[h!]\centering
\begin{minipage}{0.95\textwidth}  
\centering
\begin{tcolorbox} 
    \centering
      \small
    \begin{tabular}{p{0.95\textwidth}}
   { {\bf System message} } \\
You are given a task to match (attributes, objects) from two lists that have the same meaning. \\
    \midrule
    {\bf Prompt}  \\
        
        Inputs:\\
        
        1. ``gt-att-obj'': A dictionary with order being the key and the ground-truth (attribute, object) pair being the value. Sometimes one object can be, for example ``(black, bag), (white, bag), (striped, bag)'', it means either ``black'' or ``white'' or "striped" is correct for an attribute related with the ``bag'' and should be matched. 
        
        2. ``generated-att-obj'': A dictionary with order being the key and the generated (attribute, object) pair being the value. The order is the order of the object in the generated caption. \\\\

        Matching Criteria:
        
        - For each (attribute, object) in ``generated-att-obj'', find the (attribute, object) in the ``gt-att-obj'' that have the same meaning and add it to the ``matched-att-obj'' dictionary.
        
        - By the same meaning, we mean the words can be synonyms, can be plural/singular forms of each other and can also have different length of words to express the same meaning of attributes or objects, etc. 
        
        - If you find that the ``generated-att-obj'' can be matched with the ``gt-att-obj'' but the attribute or object in ``generated-att-obj'' is a broader concept of the attribute or object in ``gt-att-obj'', for example, one object in ``generated-att-obj'' is ``person'', but the ``gt-att-obj'' don't have ``person'' but specifically have ``man'', which is a subcategory of ``person'', add it to the ``broader-concept'' dictionary instead of the ``matched-att-obj''.\\\\
        
        Output:
        
        1. A ``broader-concept'' dictionary: \{``ORDER2'': \{``(ATTRIBUTE1, OBJECT1)'': ``(ATTRIBUTE2, OBJECT2)''\}\} only if an (ATTRIBUTE1, OBJECT1) with ORDER1 from ``generated-att-obj'' denotes a broader category of an (ATTRIBUTE2, OBJECT2) with ORDER2 in ``gt-att-obj''. Notify that Key must be the (ATTRIBUTE1, OBJECT1)from ``generated-att-obj'', Value must be (ATTRIBUTE2, OBJECT2) from ``gt-att-obj''. If none, it should be an empty dictionary. ORDER1 should be the same as ORDER2.
        
        2. A ``matched-att-obj'' dictionary: \{``ORDER2'': \{``(ATTRIBUTE1, OBJECT1)'': ``(ATTRIBUTE2, OBJECT2)''\}\} only if an (ATTRIBUTE1, OBJECT1) with ORDER1 from ``generated-att-obj'' can be mapped to an (ATTRIBUTE2, OBJECT2) with ORDER2 in ``gt-att-obj'' with the matching criteria. Key must be (ATTRIBUTE1, OBJECT1) from ``generated-att-obj'', Value must be (ATTRIBUTE2, OBJECT2) from ``gt-att-obj''. It should not contain any (ATTRIBUTE1, OBJECT1) or (ATTRIBUTE2, OBJECT2) from the ``broader-concept'' dictionary. ORDER1 should be the same as ORDER2.
        
        - The keys in ``broader-concept'' and ``matched-att-obj'' must be the same as ``gt-att-obj".\\\\
        
        For clarity, consider these examples: \textcolor[rgb]{0,0.7,0}{ \{In-context examples\} } \\
------------------- \\
        With these examples in mind, please help me extract the broader-concept, and matched-objects from the following two objects lists. \\ 
        1. gt-objects: \textcolor[rgb]{0.8,0,0}{\{Input Ground Truth Attributes\}}  \\
        2. generated-objects: \textcolor[rgb]{0.8,0,0}{\{Input Generated Attributes\}} \\\\
\\

    \end{tabular}
\end{tcolorbox}
\caption{Prompt template for matching \textbf{attributes (object)} in image caption and reference caption.
\textcolor[rgb]{0,0.7,0}{ \{In-context examples\} } are in-context examples. \textcolor[rgb]{0.8,0,0}{\{Input Ground Truth Attributes\}} are the ground truth attribute list
\textcolor[rgb]{0.8,0,0}{\{Input Generated Attributes\}} are the extracted attributes list from the extraction step which are originally captions generated by evaluated models.}
    \label{tab:matching_att_obj}
\end{minipage}
\end{table*}

\begin{table*}[h!]\centering
\begin{minipage}{0.95\textwidth}   
\centering
\begin{tcolorbox} 
    \centering

      \small
    \begin{tabular}{p{0.95\textwidth}}
   { {\bf System message} } \\
You are given a task to match (attributes, objects) from two lists that have the same meaning. \\
    \midrule
    {\bf Prompt}  \\
        
        Inputs:\\
        
        1. ``gt-att-obj'': A dictionary with order being the key and the ground-truth (attribute, object) pair being the value. Sometimes one object can be, for example ``(black, bag), (white, bag), (striped, bag)'', it means either ``black'' or ``white'' or "striped" is correct for an attribute related with the ``bag'' and should be matched. 
        
        2. ``generated-att-obj'': A dictionary with order being the key and the generated (attribute, object) pair being the value. The order is the order of the object in the generated caption. \\\\

        Matching Criteria:
        
        - For each (attribute, object) in ``generated-att-obj'', find the (attribute, object) in the ``gt-att-obj'' that have the same meaning and add it to the ``matched-att-obj'' dictionary.
        
        - By the same meaning, we mean the words can be synonyms, can be plural/singular forms of each other and can also have different length of words to express the same meaning of attributes or objects, etc. 
        
        - If you find that the ``generated-att-obj'' can be matched with the ``gt-att-obj'' but the attribute or object in ``generated-att-obj'' is a broader concept of the attribute or object in ``gt-att-obj'', for example, one object in ``generated-att-obj'' is ``person'', but the ``gt-att-obj'' don't have ``person'' but specifically have ``man'', which is a subcategory of ``person'', add it to the ``broader-concept'' dictionary instead of the ``matched-att-obj''.\\\\
        
        Output:
        
        1. A ``broader-concept'' dictionary: \{``ORDER2'': \{``(ATTRIBUTE1, OBJECT1)'': ``(ATTRIBUTE2, OBJECT2)''\}\} only if an (ATTRIBUTE1, OBJECT1) with ORDER1 from ``generated-att-obj'' denotes a broader category of an (ATTRIBUTE2, OBJECT2) with ORDER2 in ``gt-att-obj''. Notify that Key must be the (ATTRIBUTE1, OBJECT1)from ``generated-att-obj'', Value must be (ATTRIBUTE2, OBJECT2) from ``gt-att-obj''. If none, it should be an empty dictionary. ORDER1 should be the same as ORDER2.
        
        2. A ``matched-att-obj'' dictionary: \{``ORDER2'': \{``(ATTRIBUTE1, OBJECT1)'': ``(ATTRIBUTE2, OBJECT2)''\}\} only if an (ATTRIBUTE1, OBJECT1) with ORDER1 from ``generated-att-obj'' can be mapped to an (ATTRIBUTE2, OBJECT2) with ORDER2 in ``gt-att-obj'' with the matching criteria. Key must be (ATTRIBUTE1, OBJECT1) from ``generated-att-obj'', Value must be (ATTRIBUTE2, OBJECT2) from ``gt-att-obj''. It should not contain any (ATTRIBUTE1, OBJECT1) or (ATTRIBUTE2, OBJECT2) from the ``broader-concept'' dictionary. ORDER1 should be the same as ORDER2.
        
        - The keys in ``broader-concept'' and ``matched-att-obj'' must be the same as ``gt-att-obj". \\\\
        
        For clarity, consider these examples: \textcolor[rgb]{0,0.7,0}{ \{In-context examples\} } \\
------------------- \\
        With these examples in mind, please help me extract the broader-concept, and matched-objects from the following two objects lists. \\ 
        1. gt-objects: \textcolor[rgb]{0.8,0,0}{\{Input Ground Truth Attributes\}}  \\
        2. generated-objects: \textcolor[rgb]{0.8,0,0}{\{Input Generated Attributes\}} \\
\\

    \end{tabular}
\end{tcolorbox}
\caption{Prompt template for matching \textbf{attributes (people)} in image caption and reference caption.
\textcolor[rgb]{0,0.7,0}{ \{In-context examples\} } are in-context examples. \textcolor[rgb]{0.8,0,0}{\{Input Ground Truth Attributes\}} are the ground truth attribute list
\textcolor[rgb]{0.8,0,0}{\{Input Generated Attributes\}} are the extracted attributes list from the extraction step which are originally captions generated by evaluated models.}
    \label{tab:matching_att_people}
\end{minipage}
\end{table*}

\begin{table*}[h!]\centering
\begin{minipage}{0.95\textwidth}   
\centering
\begin{tcolorbox} 
    \centering
      \small
    \begin{tabular}{p{0.95\textwidth}}
   { {\bf System message} } \\
You are given a task to match (object-1 positional relation with object-2) from a ground truth dictionary and a list based on their meaning. \\
    \midrule
    {\bf Prompt}  \\
        
       Inputs:\\
        1. ``gt-relations'': A dictionary of ground truth relations. Each key is a number with no meaning of order. Each key represents different relations. The values is a list of one or two relations, 
        if there are two relations, they are synonyms. Sometimes in one relation it contains for example ``image / table'', it means either image or table in this phrase is correct. \\
        2. ``generated-relations'': A list of generated relations from a model. \\\\

        Matching Criteria:\\
        - For each relation in ``generated-relations'', find the corresponding relation in ``gt-relations'' based on their meaning, if there is none, skip it.\\
        - If you find a match, add it to the ``matched-relations'' dictionary. Note that if there are two relations in a item of ``gt-relations'', 
        it means the same meaning of the relation, you can pick either one of them as the match to the relation in ``generated-relations''.\\
        - If you find that the generated relation is a broader concept of a relation in ``gt-relations'' such as the generated relation is near each other, next to, in touch etc. \\
        but the gt-relation specifically have their relation is specifically left, right, behind or front, etc, which is more than near, add it to the ``broader-concept'' dictionary. \\\\
        
        Output: \\
        1. A ``broader-concept'' dictionary: only if an relation from ``generated-relations'' denotes a broader category of a concept in ``gt-relations'' Notify that Key must be the item from ``generated-relations'', Value must be item from ``gt-relation''. If none, it should be an empty dictionary.\\
        2. A ``matched-relations'' dictionary: only if an relation from ``generated-relations'' can be mapped to an relation in ``gt-relations'' with the matching criteria. Key must be word from ``generated-relations'', Value must be word from ``gt-relations''. It should not contain any words from the ``broader-concept'' dictionary.\\\\

        For clarity, consider these examples: \textcolor[rgb]{0,0.7,0}{ \{In-context examples\} } \\
------------------- \\
        With these examples in mind, please help me extract the broader-concept, and matched-relations from the following two inputs. \\ 
        1. gt-relations: \textcolor[rgb]{0.8,0,0}{\{Input Ground Truth Relations\}}  \\
        2. generated-relations: \textcolor[rgb]{0.8,0,0}{\{Input Generated Relations\}} \\
\\

    \end{tabular}
\end{tcolorbox}
%\vspace{-2mm}
\caption{Prompt template for matching \textbf{positional relations} in image caption and reference caption.
\textcolor[rgb]{0,0.7,0}{ \{In-context examples\} } are in-context examples. \textcolor[rgb]{0.8,0,0}{\{Input Ground Truth Relations\}} are the ground truth relation list
\textcolor[rgb]{0.8,0,0}{\{Input Generated Relations\}} are the extracted relation list from the extraction step which are originally captions generated by evaluated models.}
    \label{tab:matching_pos}
\end{minipage}
\end{table*}

\begin{table*}[h!]\centering
\begin{minipage}{0.95\textwidth}   
\centering
\begin{tcolorbox} 
    \centering

      \small
    \begin{tabular}{p{0.95\textwidth}}
   { {\bf System message} } \\
You are given a task to match the correct objects with the same meaning from a ground truth dictionary and a generated dictionary. \\
    \midrule
    {\bf Prompt}  \\
        
       Inputs:\\
        1. ``gt-objects'': A dictionary of ground truth objects. Each key is a number starting rank No.1 and increment each time by 1. Each value is the corresponding object with the rank. 
        Sometimes one object can be, for example ``ground / court'', it means either ground or court is correct and should be matched.  \\
        2. ``generated-objects'': A dictionary with rank being the key and the object being the value. The rank is the rank of the object in the generated caption.\\\\

        Matching Criteria:\\
        - For each object in ``generated-objects'', find the object in the ``gt-objects'' that have the same meaning and add it to the ``matched-objects'' dictionary. \\
        - By the same meaning, we mean the words can be synonyms, can be plural/singular forms of each other and can also have different length of words to express the same meaning of objects, etc. \\
        - Notice that the final matched-objects must follow the order of values in ``generated-objects''. \\
        - If you find that the ``generated-objects'' can be matched with the ``gt-objects'' but the object in ``generated-objects'' is a broader concept of the objects in ``gt-objects'', for example, one object in ``generated-objects'' is ``person'', but the ``gt-objects'' don't have ``person'' but specifically have ``man'', which is a subcategory of ``person'', add it to the `broader-concept'' dictionary instead of the ``matched-objects''.\\\\
        
        Output: \\
        1. A ``broader-concept'' dictionary: only if an object from ``generated-objects'' denotes a broader category of a concept in ``gt-objects'' Notify that Key must be the item from ``generated-objects'', Value must be item from ``gt-objects''. If none, it should be an empty dictionary. \\
        2. A ``matched-objects'' dictionary: only if an object from ``generated-objects'' can be mapped to an object in ``gt-objects'' with the matching criteria. Key must be word from ``generated-objects'', Value must be word from ``gt-objects''. It should not contain any words from the ``broader-concept'' dictionary.\\\\

        For clarity, consider these examples: \textcolor[rgb]{0,0.7,0}{ \{In-context examples\} } \\
------------------- \\
        With these examples in mind, please help me extract the broader-concept, and matched-relations from the following two inputs. \\ 
        - gt-relations: \textcolor[rgb]{0.8,0,0}{\{Input Ground Truth Relations\}}  \\
        - generated-relations: \textcolor[rgb]{0.8,0,0}{\{Input Generated Relations\}} \\
\\

    \end{tabular}
\end{tcolorbox}
%\vspace{-2mm}
\caption{Prompt template for matching \textbf{comparative relations} in image caption and reference caption.
\textcolor[rgb]{0,0.7,0}{ \{In-context examples\} } are in-context examples. \textcolor[rgb]{0.8,0,0}{\{Input Ground Truth Relations\}} are the ground truth objects ranking list
\textcolor[rgb]{0.8,0,0}{\{Input Generated Relations\}} are the extracted objects list from the extraction step which are originally captions generated by evaluated models.}
    \label{tab:matching_comparative}
\end{minipage}
\end{table*}

\section{Qualitative Results}
\label{apx:qualitative_results}

\begin{figure*}[ht!]
 \centering
 \includegraphics[width=\textwidth]{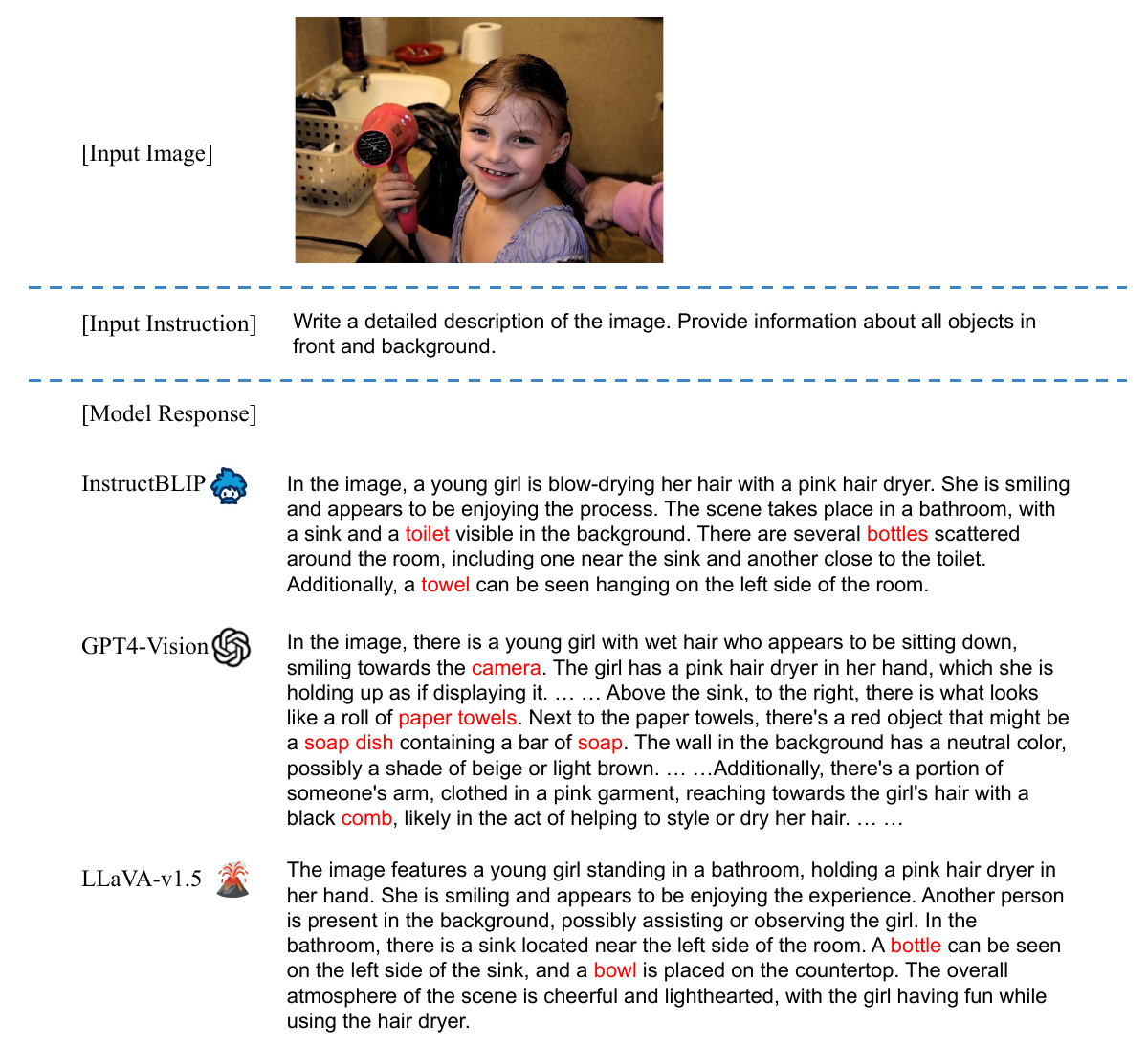}
 \caption{Object existence evaluation example from three representative models in our benchmark \data~. Text in red indicating models' hallucinations.}
 \label{fig:example_object}
\end{figure*}

\begin{figure*}[ht!]
 \centering
 \includegraphics[width=\textwidth]{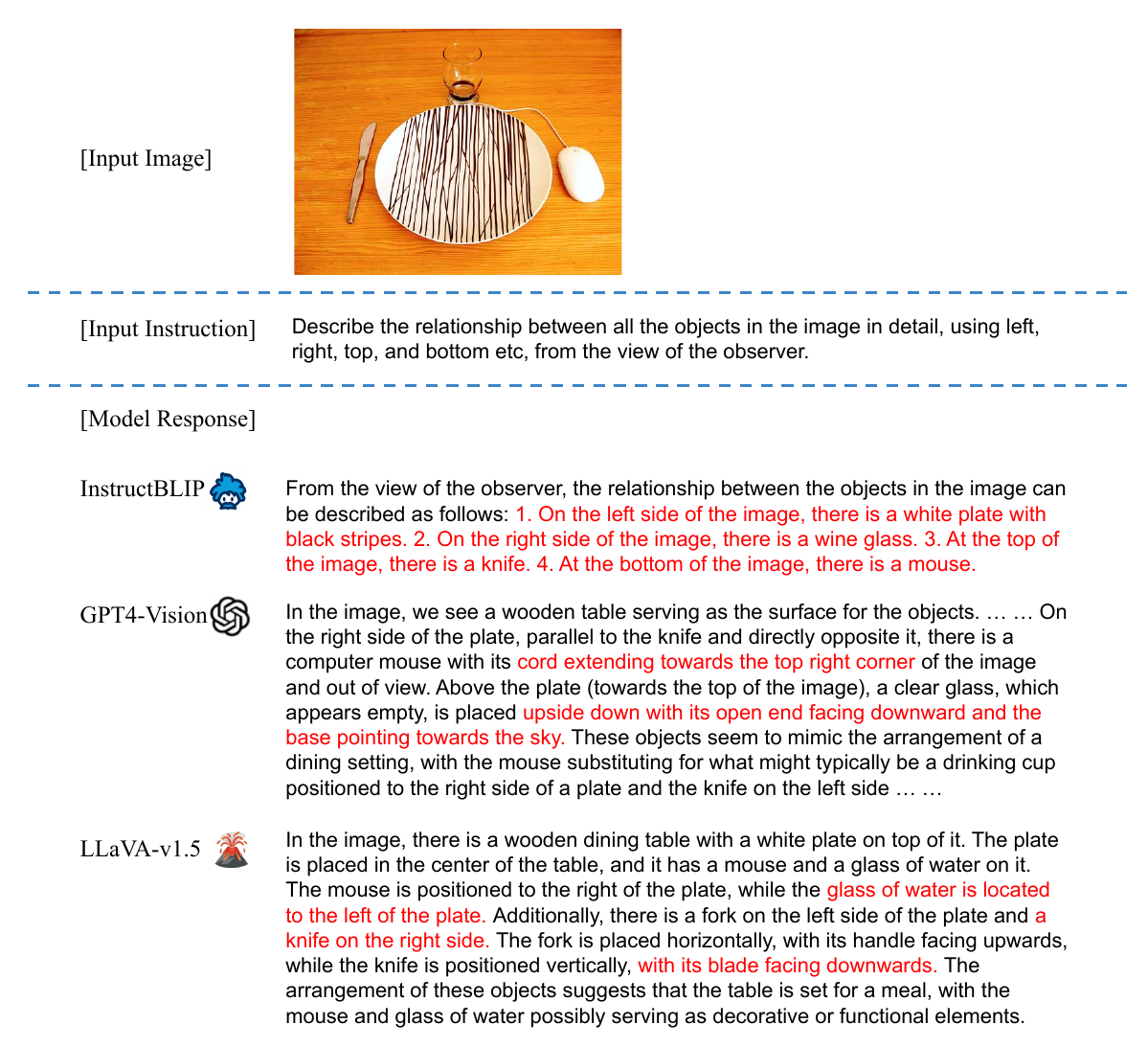}
 \caption{Positional relation evaluation example from three representative models in our benchmark \data~. Text in red indicating models' hallucinations.}
 \label{fig:example_pos_rela}
\end{figure*}

\begin{figure*}[ht!]
 \centering
 \includegraphics[width=\textwidth]{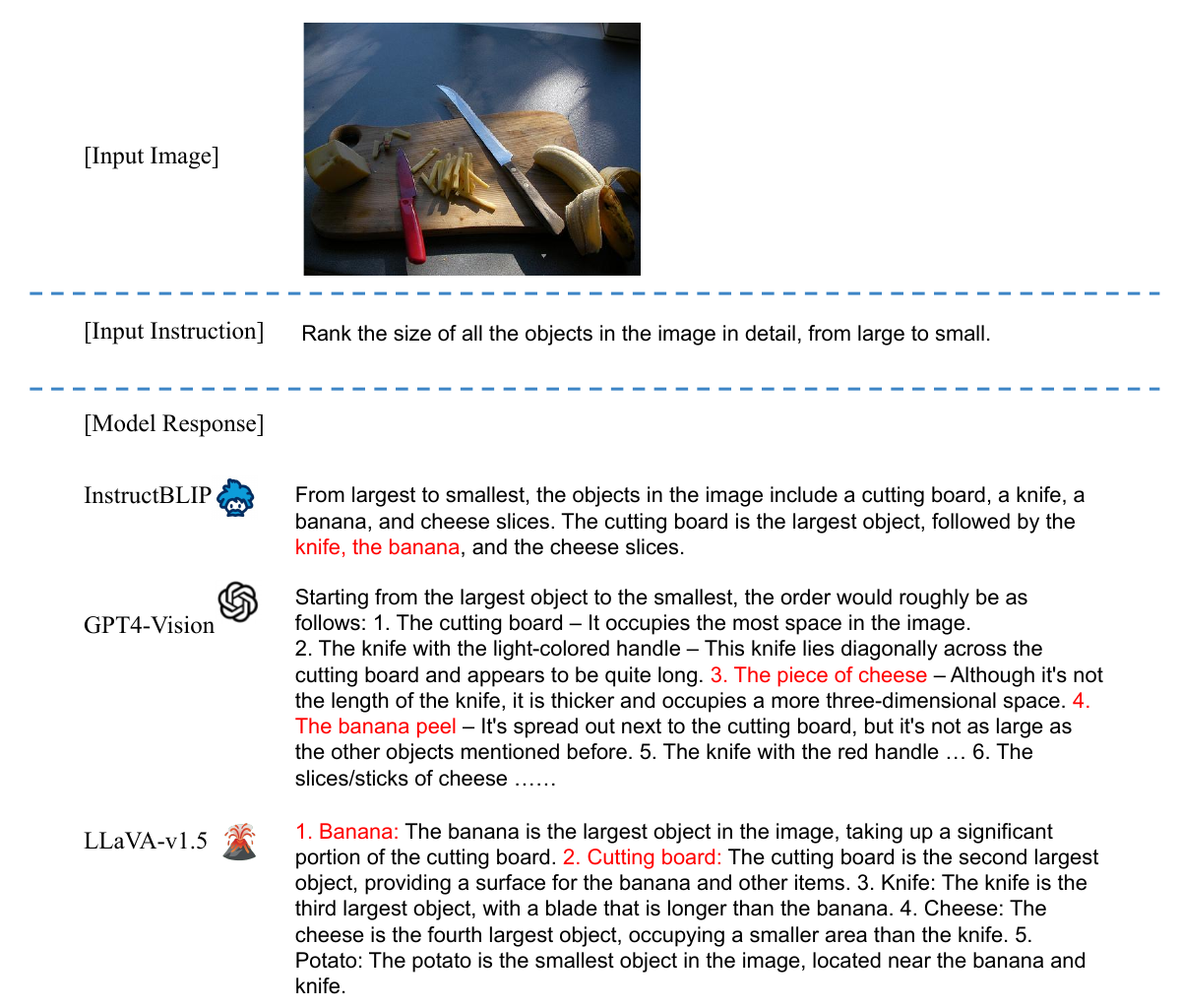}
 \caption{Comparative relation evaluation example from three representative models in our benchmark \data~. Text in red indicating models' hallucinations.}
 \label{fig:example_comp_rela}
\end{figure*}

We illustrate the qualitative results of three representative models in \Cref{fig:example_object}, \Cref{fig:example_pos_rela} and \Cref{fig:example_comp_rela}. Each model exhibited instances of hallucination in these examples from our benchmark \data~. Notably, while GPT-4V generates the most comprehensive results, it is also more prone to producing hallucinations.

\end{document}